\documentclass{article}

     \usepackage[preprint]{neurips_2019}

\usepackage[utf8]{inputenc} 
\usepackage[T1]{fontenc}    
\usepackage{hyperref}       
\usepackage{url}            
\usepackage{booktabs}       
\usepackage{amsfonts,amssymb,amsthm,amsmath,latexsym}     
\usepackage{nicefrac}       
\usepackage{microtype}      
\theoremstyle{plain}

\title{Reinforcement Learning is not a Causal problem}

\author{
  Mauricio Gonzalez-Soto \\
  Coordinación de Ciencias Computacionales\\
  Instituto Nacional de Astrofisica Optica y Electronica (INAOE)\\
  Mexico \\
  \texttt{mauricio@inaoep.mx}
   \And
  Felipe Orihuela Espina\\
   Coordinación de Ciencias Computacionales \\
   Instituto Nacional de Astrofisica Optica y Electronica (INAOE) \\
   Mexico\\
   \texttt{f.orihuela-espina@inaoep.mx}
}

\begin{document}

\maketitle
\begin{abstract}
We use an analogy between non-isomorphic mathematical structures defined over the same set and the algebras induced by associative and causal levels of information in order to argue that Reinforcement Learning, in its current formulation, is not a causal problem, independently if the motivation behind it has to do with an agent taking actions. 
\end{abstract}
\section{Introduction}
There is an open debate about Reinforcement Learning (RL) being a Causal problem or not. According to \cite{sutton1998reinforcement}, the RL problem is to learn some task in an interactive way, and the now standard solution consists in assigning values for the different states, or values, which certain stochastic proces can take. It has been argued that RL is essentially a control problem, and since in a RL problem agent performs interventions in a given environment, then this is a causal problem (\cite{csaba}, \cite{sutton1992reinforcement}). RL can also help to understand human-learning processes and their causal interpretation (\cite{gershman2015reinforcement}); in fact, since human beings conceive their actions as interventions in the world, and humans actively perceive their environment by predicting the outcomes of such interventions (\cite{clark2015surfing}) it is tempting to consider RL as a causal problem in nature. 

The objective of the most-common used RL algorithms is to find a \textit{policy}, which is a map between states and actions, which is interpreted as what should a rational agent do if he finds himself in such state. RL, both in its formulation, its optimality criteria and the algorithms involved, use operations based on associative information, but do not make use of causal operations. Here we argue that RL and causal reasoning are inherently different problems, and by presenting an analogy in terms of algebraic structures we argue that once established the different algebras that associative and causal information induce we can not mix between them irrespective of the \textit{motivation} or real-life situation that lies behind.
\section{Levels of formulation of a problem}
When talking about \textit{a problem} one must be careful and distinguish between a \textit{real world situation} and what \textit{mathematical formulation} of such situation. In the problem of learning by interaction, the intuition is of an intelligent agent manipulating his environment and learning from the consequences of his actions via a reward function. The standard formulation of such problems is through a Markov Decision Process, or some variants of it. An \textit{optimal policy} is what the scientific comunity has accepted to be the solution of mathematical problem generated the learning by interaction problem, and several algorithms have been proposed to find a such policy.

Even while the intuition behind RL is that of an agent \textit{interacting} with an environment, it does not mean that the mathematical model of such agent captures the notion of his actions as \textit{interventions} in the environment; this only remains from a linguistic confusion between a real life situation, and a mathematical model. RL and the mathematical tools used in its formulation operate only at the associative level of information; this is, RL can only learn from correlations in data. As Pearl puts it, RL only operates in the first level of causal reasoning and lacks the necessary tools of the upper levels: interventions and counterfactuals. 

\section{Causal and associative algebras}
As a simpler case, consider the structures $G_1 = (\mathbb{Z}, +)$ where $+$ is the usual sum, and $ G_2 = (\mathbb{Z}, \cdot)$ where $\cdot$ is the usual multiplication. It is clear that $G_1$ and $G_2$ do not have the same algebraic structure; more specificaly, $G_1$ is an abelian group while $G_2$ is only a semigroup (\cite{hungerford}); therefore, any equation stated in $G_1$ can not be solved using methods valid for $G_2$. This is, consider the equation:
\begin{equation}
a+b=c,
\end{equation}
which must be solved for $a$ if $b$ and $c$ are known. Given that $G_1$ and $G_2$ are clearly not isomorphic, we can not attempt to solve for $a$ using any insight provided by knowledge of $G_2$; even if, on an upper level, we knew that on $(\mathbb{R}, +, \cdot)$ $b$ has the form, say, $b=d^f$, we must solve for $a$ only in the domain of addition; other examples are the $\mathbb{Z}_4$ group, which is not isomorphic to Klein's group $\mathbb{Z}_2 \times \mathbb{Z}_2$, or the Hamilton quaternions $\mathbb{H}$, which are an abelian group under addition but not under its respective multiplication. 

Considering the manipulationist notion of causation (\cite{woodward2005making}), which contains both Pearl's Structural Causal Models (\cite{pearl2009causality}) and Spirtes' Causation (\cite{spirtes2000causation}), we recognize two fundamental aspects: an implicit order $\succeq$ and the presence of a context $Z$. On the other hand, if considering only associative information, there is no distinguishable order even in the case of stochastically dependant variables; notice that any distribution $p(X,Y)$ can be expressed either as $p(X | Y)p(Y)$ or $P(Y|X)p(Y)$.

Let $\bigoplus$ and $\bigotimes$ the operators representing the associative algebra and the causal algebra; i.e., two variables $A$, and $B$ which are correlated are represented as $\bigoplus(A,B)$ while a variable, or event $A$ which causes some other $B$ are represented by $\bigotimes(A,B)$. Even more specificaly, $\bigotimes$ should be written as $\bigotimes(Z, \succeq)$

\section{Reinforcement Learning}
Let 
\begin{equation}
a R(\beta) b
\end{equation}
a relation between $a$ and $b$, where $\beta$ are parameters which express $R$ univocally. In particular, in RL we must find
\begin{equation}
\beta = f[a R b]
\end{equation}
where $R= \bigoplus$, and $f$ a function which depends on the state and reward of the system through only associative operations (e.g., the $Q$ function). And here in this point we have our main argument: since $R$ is the associative algebra, and such algebra can not be isomortphic to the causal algebra because of the lack of order, then we can not use causal tools to solve for $\beta$, and therefore RL is not a causal problem. This is,
\begin{equation}
a \bigoplus b
\end{equation}
and
\begin{equation}
a \bigotimes(Z, \succeq) b
\end{equation}
are different problems, which must be solved with their respective tools. This said, current reinforcement learning problems can not be considered to be causal if their mathematical formulation relies only on associative tools.
\section{Conclusion}
We have argued that problems that can be solved at the associative level of information must be solved using the respective tools, and the same applies for causal problems. One must be careful not to mix the language and framework induced by the chosen formulation in order to model some real-life situation. The classical RL formulation could, in principle, be modified in order to allow a  proper causal formulation; we speculate that the Bellman equations for the $V$ and $Q$ functions could be modified, for a deterministic policy and reward, in the following way:
\begin{equation}
v_{\pi}(s) = \sum_{s',r}\mathbb{P}^g(s' |s, do(a)) [r + \gamma v_\pi (s') ],
\end{equation}
\begin{equation}
q_\pi (s,a) = \sum_{s',r} \mathbb{P}^g(s' |s, do(a))[r + \gamma \max_{a} q(s',a')],
\end{equation}
where $\mathbb{P}^g$ is the probability distribution induced by a Causal Graphical Model $g$

\bibliographystyle{apalike}
\bibliography{/Users/MauricioGS1/INAOE/Propuesta/Bibliografia.bib}

\begin{thebibliography}{}

\bibitem[Clark, 2015]{clark2015surfing}
Clark, A. (2015).
\newblock {\em Surfing uncertainty: Prediction, action, and the embodied mind}.
\newblock Oxford University Press.

\bibitem[Gershman, 2015]{gershman2015reinforcement}
Gershman, S.~J. (2015).
\newblock Reinforcement learning and causal models.
\newblock In {\em The Oxford Handbook of Causal Reasoning}.

\bibitem[Hungerford, 1974]{hungerford}
Hungerford, T.~W. (1974).
\newblock {\em Algebra}.
\newblock Springer-Verlag New York.

\bibitem[Pearl, 2009]{pearl2009causality}
Pearl, J. (2009).
\newblock {\em Causality: Models, Reasoning and Inference}.
\newblock Cambridge University Press, New York, NY, USA, 2nd edition.

\bibitem[Spirtes et~al., 2000]{spirtes2000causation}
Spirtes, P., Glymour, C.~N., and Scheines, R. (2000).
\newblock {\em Causation, prediction and search}.
\newblock MIT Press.

\bibitem[Sutton and Barto, 1998]{sutton1998reinforcement}
Sutton, R.~S. and Barto, A.~G. (1998).
\newblock {\em Reinforcement Learning: An introduction}.
\newblock MIT Press.

\bibitem[Sutton et~al., 1992]{sutton1992reinforcement}
Sutton, R.~S., Barto, A.~G., and Williams, R.~J. (1992).
\newblock Reinforcement learning is direct adaptive optimal control.
\newblock {\em IEEE Control Systems Magazine}, 12(2):19--22.

\bibitem[Szepesvari, 2018]{csaba}
Szepesvari, C. (2018).
\newblock Causality from the perspective of reinforcement learning.
\newblock Machine Learning for Causal Inference, Counterfactual Prediction, and
  Autonomous Action (CausalML) Workshop, ICML.

\bibitem[Woodward, 2003]{woodward2005making}
Woodward, J. (2003).
\newblock {\em Making things happen: A theory of causal explanation}.
\newblock Oxford Studies in Philosophy of Science. Oxford University Press.

\end{thebibliography}
\end{document}